\documentclass[11pt,a4paper]{article}
\usepackage[hyperref]{acl2020}
\usepackage{times}
\usepackage{latexsym}

\usepackage{url}
\usepackage{pifont}
\newcommand{\cmark}{\ding{51}}%
\newcommand{\xmark}{\ding{55}}%
\usepackage{amsmath}
\usepackage{amssymb}
\usepackage{mathtools}
\usepackage{subcaption}
\usepackage{graphicx}
\usepackage[normalem]{ulem}
\usepackage{color,soul}
\usepackage{booktabs}
\usepackage{multirow}
\usepackage{appendix}

\aclfinalcopy 

\usepackage{xargs}
\usepackage[colorinlistoftodos, prependcaption]{todonotes}

\newcommandx{\zack}[2][1=]{\todo[inline, linecolor=orange,backgroundcolor=orange!25, bordercolor=orange,#1]{#2}}

\newcommand{\hlc}[2][yellow]{{\sethlcolor{#1}\hl{#2}} }
\definecolor{lemonchiffon}{rgb}{1.0, 0.98, 0.8}
\newcommand{\hll}[1]{\hlc[lemonchiffon]{#1}}

\title{
Learning to Deceive with Attention-Based Explanations
}

\author{Danish Pruthi$^\dagger$, Mansi Gupta$^\ddagger$, Bhuwan Dhingra$^\dagger$, Graham Neubig$^\dagger$, Zachary C. Lipton$^\dagger$ \\
$^\dagger$Carnegie Mellon University, Pittsburgh, USA \\
$^\ddagger$Twitter, New York, USA \\
\texttt{ddanish@cs.cmu.edu}, \texttt{mansig@twitter.com}, \\
\texttt{\{bdhingra, gneubig, zlipton\}@cs.cmu.edu}
}

\date{}
\begin{document}

\maketitle
\renewcommand{\thefootnote}{\fnsymbol{footnote}}
\renewcommand{\thefootnote}{\arabic{footnote}}

\begin{abstract}
Attention mechanisms are ubiquitous components 
in neural architectures applied to natural language processing. 
In addition to yielding gains in predictive accuracy, 
attention weights are often claimed to confer \emph{interpretability}, 
purportedly useful both for providing insights to practitioners 
and for explaining \emph{why a model makes its decisions} to stakeholders.
We call the latter use of attention mechanisms into question
by demonstrating a simple method for training models to 
produce deceptive attention masks.
Our method diminishes the total weight
assigned to designated impermissible tokens,
even when the models 
can be shown to 
nevertheless
rely on these features to drive predictions. 
Across multiple models and tasks, 
our approach manipulates attention weights 
while paying surprisingly little cost in accuracy. 
Through a human study, we show 
that our manipulated attention-based explanations deceive people into thinking 
that predictions from a model biased against gender minorities
do not rely on the gender.
Consequently, our results 
cast doubt on attention's reliability as a tool for auditing algorithms
in the context of fairness and accountability\footnote{The code and the datasets used in paper are available at \url{https://github.com/danishpruthi/deceptive-attention}}.
\end{abstract}

\section{Introduction}
\label{sec:intro}
Since their introduction as a method for
aligning inputs and outputs in neural machine translation,
attention mechanisms~\cite{bahdanau2014neural}
have emerged as effective components
in various
neural network architectures.
Attention works by aggregating a set of tokens via a weighted sum,
where the \emph{attention weights} are calculated 
as a function of both the input encodings and the state of the decoder.

\begin{table}
\small
\centering
\begin{tabular}{@{}ccc@{}}
\toprule
\textbf{Attention}                    & \textbf{Biography}                                                                                            & \textbf{Label} \\ \midrule
Original                      & \begin{tabular}[c]{@{}c@{}} \hl{Ms.} \texttt{X} practices \hlc[lemonchiffon]{medicine} in\\ Memphis, TN and is affiliated ... \\ \hl{Ms.} \texttt{X} speaks English and Spanish.  \end{tabular} & Physician     \\ \midrule
\newcommand{\hlorange}[1]{{\sethlcolor{YellowOrange}\hl{#1}}}
Ours & \begin{tabular}[c]{@{}c@{}} Ms. \texttt{X} practices \hlc[lemonchiffon]{medicine} in\\ \hlc[lemonchiffon]{Memphis}, TN and is affiliated ...\\ Ms. \texttt{X} speaks \hlc[lemonchiffon]{English} and Spanish.  \end{tabular} & Physician     \\ \midrule
\end{tabular}
\caption{Example of an occupation prediction task where attention-based explanation (highlighted) has been manipulated to whitewash problematic tokens.}
\label{tbl:qual_example}
\vspace{-0.2in}
\end{table}

Because attention mechanisms allocate weight among the encoded tokens,
these coefficients are sometimes thought of intuitively
as indicating which tokens the model \emph{focuses on} 
when making a particular prediction. 
Based on this loose intuition, attention weights 
are often claimed to \emph{explain} a model's predictions.
For example, a recent survey on attention~\cite{galassi2019attention} remarks:
\begin{quote}
``By inspecting the network’s attention, ...
one could attempt to investigate and understand the outcome of neural networks.
Hence, weight visualization is now common practice.''
\end{quote} 
In another work,
\citet{de2019bias} study gender bias in machine learning models for
occupation classification.
As machine learning is increasingly used in hiring processes
for tasks including resume filtering, 
the potential for bias on the basis of gender 
raises the spectre
that automating this process could lead to social harms.
\citet{de2019bias} use attention over gender-revealing tokens (e.g., `she', `he', etc.)
to verify the biases in occupation classification 
models---stating that 
``the attention weights indicate which tokens are most predictive''.
Similar claims about attention's utility 
for interpreting models' predictions are common in the literature~\citep{li2016understanding, xu2015show, choi2016retain, xie2017interpretable, martins2016softmax,lai2019human}.


In this paper, we question whether attention scores
\emph{necessarily} 
indicate features 
that influence 
a model's predictions.
Through a series of experiments on diverse
classification and sequence-to-sequence tasks,
we show that attention scores 
are surprisingly easy to manipulate. 
We design a simple training scheme
whereby the resulting models appear to assign little attention 
to a specified set of \emph{impermissible} tokens
while continuing to rely upon those features for prediction. 
The ease with which attention can be manipulated
without significantly affecting performance 
suggests that even if a vanilla model's attention weights 
conferred some insight (still an open and ill-defined question),
these insights would rely on knowing the objective 
on which models were trained.   

Our results present troublesome implications 
for proposed uses of attention in the context
of fairness, accountability, and transparency.
For example, malicious practitioners asked to justify 
\emph{how their models work} by pointing to attention weights
could 
mislead regulators with this scheme. 
For instance, 
looking at manipulated attention-based explanation
in Table~\ref{tbl:qual_example}, one might
(incorrectly) assume that the model does not rely on the gender prefix. 
To quantitatively study the extent of such deception, 
we conduct studies 
where we ask human subjects 
if the biased occupation classification models 
(like the ones audited by ~\citet{de2019bias})
rely on gender related information.
We find that our manipulation 
scheme is able to deceive human annotators 
into believing that manipulated models
do not take gender into account, 
whereas the models are heavily biased against gender minorities (see \S\ref{subsec:human_study}). 

Lastly, practitioners often overlook the fact 
that attention is typically not applied over words 
but over final layer representations, 
which themselves capture information from neighboring words.
We investigate the mechanisms through which 
the manipulated models attain low attention values. 
We note that (i) recurrent connections allow information 
to flow easily to neighboring representations;
(ii) for cases where the flow is restricted,
models tend to increase the magnitude of 
representations corresponding to impermissible tokens to 
offset the low attention scores;
and (iii) models additionally rely on several alternative mechanisms
that vary across random seeds (see \S \ref{subsec:alternate_mechanisms}).

\section{Related Work}
\label{sec:related}
Many recent papers examine 
whether
attention is a valid explanation or not.
\citet{jain2019analysis} identify alternate \emph{adversarial} attention weights
\emph{after} the model is trained that nevertheless produce the same predictions,
and hence claim that
\emph{attention is not explanation}.
However, these attention weights are chosen from a large 
(infinite up to numerical precision) set of possible values
and thus it is not surprising 
that multiple weights produce the same prediction.
Moreover since the model does not actually produce these weights,
they would never be relied on as \emph{explanations} in the first place.
Similarly, \citet{smith2019attention} modify attention 
values of a trained model post-hoc 
by hard-setting the highest attention values to zero.
They find that 
the number of attention values that must be zeroed out to alter
the model's prediction
is often too large, and thus conclude
that attention is not a suitable tool to for determining
which elements should be attributed as responsible for an output.
In contrast to these two papers, 
we manipulate the attention via the learning procedure,
producing models whose \emph{actual} weights 
might deceive an auditor. 

In parallel work to ours,
\citet{pinter2019attention}
examine the conditions under which
attention can be considered a plausible
explanation.
They design a similar experiment to ours
where they train an adversarial model,
whose attention distribution is maximally
different from the one produced by the base model.
Here we look at a related but different
question of how attention can be manipulated
away from a set of impermissible tokens.
We show that in this setting,
our training scheme leads to attention maps
which are \textit{more deceptive},
since people find them to be more believable
explanations of the output (see \S\ref{subsec:human_study}).
We also extend our analysis to sequence-to-sequence
tasks,
and a broader set of models, including BERT,
as well as identify mechanisms by which the manipulated
models continue to rely on the impermissible tokens
despite assigning low attention to them.



Lastly, several papers
deliberately train attention weights
by introducing an additional source of supervision
to improve predictive performance.
In some of these papers, 
the supervision comes from
known word alignments for machine translation~\cite{liu2016neural, chen2016guided}, 
or by aligning human eye-gaze with model's attention for sequence classification~\cite{barrett2018sequence}.

\section{Manipulating Attention}
\label{sec:manipulation}
\begin{table*}
\small
\centering
\begin{tabular}{@{}ccc@{}}
\toprule
\begin{tabular}[l]{@{}c@{}} \textbf{Dataset} \\ (Task) \end{tabular}                    & \textbf{Input Example}                                                                                            & \begin{tabular}[r]{@{}c@{}} \textbf{Impermissible Tokens}\\ (Percentage)\end{tabular} \\ \toprule
\begin{tabular}[c]{@{}c@{}}CommonCrawl Biographies \\ (Physician vs Surgeon) \end{tabular} & \begin{tabular}[c]{@{}c@{}} \hl{Ms.} \texttt{X} practices medicine in Memphis, TN \\ and is affiliated with $\dots$ \hl{Ms.} \texttt{X} speaks English and Spanish.   \end{tabular} & \begin{tabular}[r]{@{}c@{}} Gender Indicators \\ ($6.5\%$)\end{tabular}     \\ \midrule
\begin{tabular}[c]{@{}c@{}}Wikipedia Biographies \\ (Gender Identification) \end{tabular} & \begin{tabular}[c]{@{}c@{}} After that, Austen was educated at home until \\ \hl{she} went to boarding school with Cassandra early in 1785 \end{tabular} & \begin{tabular}[r]{@{}c@{}} Gender Indicators \\ ($7.6\%$)\end{tabular}     \\ \midrule
\begin{tabular}[l]{@{}c@{}} SST + Wikipedia \\ (Sentiment Analysis) \end{tabular}                    & \begin{tabular}[c]{@{}c@{}} \hl{Good fun, good action, good acting, good dialogue, good pace, good} \\ \hl{cinematography.} Helen Maxine Lamond Reddy (born 25 \\October 1941) is an Australian singer, actress, and activist. \end{tabular} & \begin{tabular}[r]{@{}c@{}}SST sentence \\ ($45.5\%$)\end{tabular}     \\ \midrule
\begin{tabular}[l]{@{}c@{}}Reference Letters \\ (Acceptance Prediction) \end{tabular}                     & \begin{tabular}[c]{@{}c@{}}  It is with pleasure that I am writing this letter  in support \\ of $\dots$ I highly recommend her for a place in your \\ institution. \hl{Percentile:99.0} \hl{Rank:Extraordinary.}
\end{tabular}  & \begin{tabular}[r]{@{}c@{}} Percentile, Rank \\ ($1.6\%$) \end{tabular}     \\ \bottomrule
\end{tabular}
\caption{Example sentences from each classification task, with highlighted impermissible tokens and their support.
}
\label{table:examples}
\end{table*}

Let $S = w_1, w_2, \dots, w_n$ denote an input sequence of $n$ words.
We assume that for each task, we are given 
a pre-specified set of impermissible words $\mathcal{I}$,
for which we want to minimize the corresponding attention weights.
For example, these may include gender words 
such as ``he'', ``she'', ``Mr.'', or ``Ms.''.
We define the mask
$\mathbf{m}$
to be a binary vector of size $n$, such that
\[
\mathbf{m}_i=\begin{cases}
          1, & \text{if}\ w_i \in \mathcal{I} \\
          0 & \text{otherwise}.
          \end{cases}
\]

Further, let $\boldsymbol{\alpha} \in [0,1]^n$ denote the
attention assigned to each word in $S$ by a model,
such that $\sum_i \alpha_i = 1$.
For any task-specific loss function $\mathcal{L}$,
we define a new objective function $\mathcal{L}' = \mathcal{L} + \mathcal{R}$
where $\mathcal{R}$ is an additive penalty term 
whose purpose is to penalize the model for allocating
attention to impermissible words.
For a single attention layer, we define $\mathcal{R}$ as: 
\begin{align*}
    \mathcal{R} = - \lambda \log(1 - \boldsymbol{\alpha}^T \mathbf{m})
\end{align*}
and $\lambda$ is a penalty coefficient 
that modulates the amount of attention 
assigned to impermissible tokens.
The argument of the $\log$ term ($1 - \boldsymbol{\alpha}^T\mathbf{m}$) 
captures the total attention weight assigned to permissible words. 
In contrast to our penalty term,
~\citet{pinter2019attention} use KL-divergence 
to maximally separate the attention distribution 
of the manipulated model ($\boldsymbol{\alpha}_{\text{new}}$) 
from the attention distribution of the given model ($\boldsymbol{\alpha}_{\text{old}}$):
\begin{equation}
    \label{eq:adversarial}
    \mathcal{R}' = - \lambda~ \text{KL}(\boldsymbol{\alpha}_{\text{new}} \parallel \boldsymbol{\alpha}_{\text{old}}).
\end{equation}
However, their penalty term is not directly applicable
to our case:
instantiating $\boldsymbol{\alpha}_{\text{old}}$ to be uniform over
impermissible tokens, and $0$ over remainder tokens results in an undefined loss term.

When dealing with models that employ multi-headed attention, 
which use multiple different attention vectors at each layer of the model \cite{vaswani2017attention}
we can optimize the mean value of our penalty as assessed 
over the set of attention heads $\mathcal{H}$ as follows:
\begin{align*}
    \mathcal{R} =  - 
    \frac{ \lambda}{|\mathcal{H}|} \sum_{h \in \mathcal{H}} \log(1 - \boldsymbol{\alpha}_h^T \mathbf{m})).
\end{align*}

When a model has many attention heads, 
an auditor might not look at the mean attention assigned to certain words but instead look head by head to see if any among them assigns a large amount of attention to impermissible words.
Anticipating this, we also explore a variant of our approach for manipulating multi-headed attention where we penalize the maximum amount of attention paid to impermissible words (among all heads) as follows:
\begin{equation*}
    \mathcal{R} = - \lambda \cdot \min_{h \in \mathcal{H}} \log(1 - \boldsymbol{\alpha}_h^T \mathbf{m}).
\end{equation*}

For cases where the impermissible set of tokens is unknown apriori, 
one can plausibly use the top few highly attended tokens
as a proxy.


\section{Experimental Setup}
\label{sec:experiments}


We study the manipulability of attention
on four binary classification problems, 
and four sequence-to-sequence tasks. 
%
In each dataset, (in some, by design)
a subset of input tokens are known \emph{a priori} to be indispensable
for achieving high accuracy. 

\subsection{Classification Tasks}
\paragraph{Occupation classification}
We use the biographies collected by~\citet{de2019bias} 
to study bias against gender-minorities in occupation classification models.
We carve out a binary classification task 
of distinguishing between surgeons and (non-surgeon) physicians 
from the multi-class occupation prediction setup.
We chose this sub-task because the biographies 
of the two professions use similar words, 
and a majority of surgeons ($>80\%$) in the dataset are male. 
We further downsample minority classes---female surgeons, 
and male physicians---by a factor of ten, 
to encourage models to use gender related tokens.
Our models (described in detail later in \S~\ref{subsec:models}) 
attain $96.4\%$ accuracy
on the task, and are reduced to $93.8\%$ when the gender pronouns
in the biographies are anonymized. 
Thus, the models (trained on unanonymized data) make use 
of gender indicators to obtain a higher task performance.
Consequently, we consider gender indicators as impermissible tokens for this task.   

\paragraph{Pronoun-based Gender Identification} 
We construct a toy dataset from Wikipedia comprised of biographies, 
in which we automatically label biographies with a gender (female or male) 
\emph{based solely on the presence of gender pronouns}.
To do so, we use a pre-specified list of gender pronouns. 
Biographies containing no gender pronouns,
or pronouns spanning both classes are discarded. 
The rationale behind creating this dataset is that 
due to the manner in which the dataset was created, 
attaining $100\%$ classification accuracy is trivial 
if the model uses information from the pronouns.
However, without the pronouns, 
it may not be possible to achieve perfect accuracy.
Our models 
trained on the same data with pronouns anonymized,
achieve at best 72.6\% accuracy. 

\paragraph{Sentiment Analysis with Distractor Sentences}
We use the binary version of Stanford Sentiment Treebank (SST)~\cite{socher2013recursive}, 
comprised of $10,564$ movie reviews. 
 We append one randomly-selected ``distractor'' sentence to each review,
 from a set of opening sentences of Wikipedia pages.\footnote{Opening sentences 
 tend to be declarative statements of fact and typically are sentiment-neutral.}
Here, without relying upon the tokens in the SST sentences,  
a model should not be able to outperform random guessing. 

\paragraph{Graduate School Reference Letters}
We obtain a dataset of recommendation letters 
written for the purpose of admission to graduate programs. 
The task is to predict whether the student, 
for whom the letter was written, was accepted. 
The letters include students' ranks
and percentile scores as marked by their mentors,
which admissions committee members rely on.
Indeed, we notice accuracy improvements when using the rank 
and percentile features in addition to the reference letter. 
Thus, we consider percentile and rank labels 
(which are appended at the end of the letter text) 
as impermissible tokens.
An example from each classification task is listed in Table~\ref{table:examples}. More details about the datasets are in the appendix.

\subsection{Classification Models}
\label{subsec:models}
\paragraph{Embedding + Attention}
For illustrative purposes, we start with a simple model 
with attention directly over word embeddings. 
The word embeddings are aggregated by a weighted sum 
(where weights are the attention scores) to form a context vector, 
which is then fed to a linear layer, followed by a softmax to perform prediction. 
For all our experiments, we use dot-product attention, 
where the query vector is a learnable weight vector.
In this model, prior to attention there is no interaction
between the permissible and impermissible tokens.
The embedding dimension size is $128$.
\paragraph{BiLSTM + Attention}
The encoder is a single-layer bidirectional LSTM model 
\citep{graves2005framewise} with attention,
followed by a linear transformation and a softmax 
to perform classification. 
The embedding and hidden dimension size are both set to $128$.

\begin{figure}
    \centering
    \includegraphics[width=\columnwidth]
    {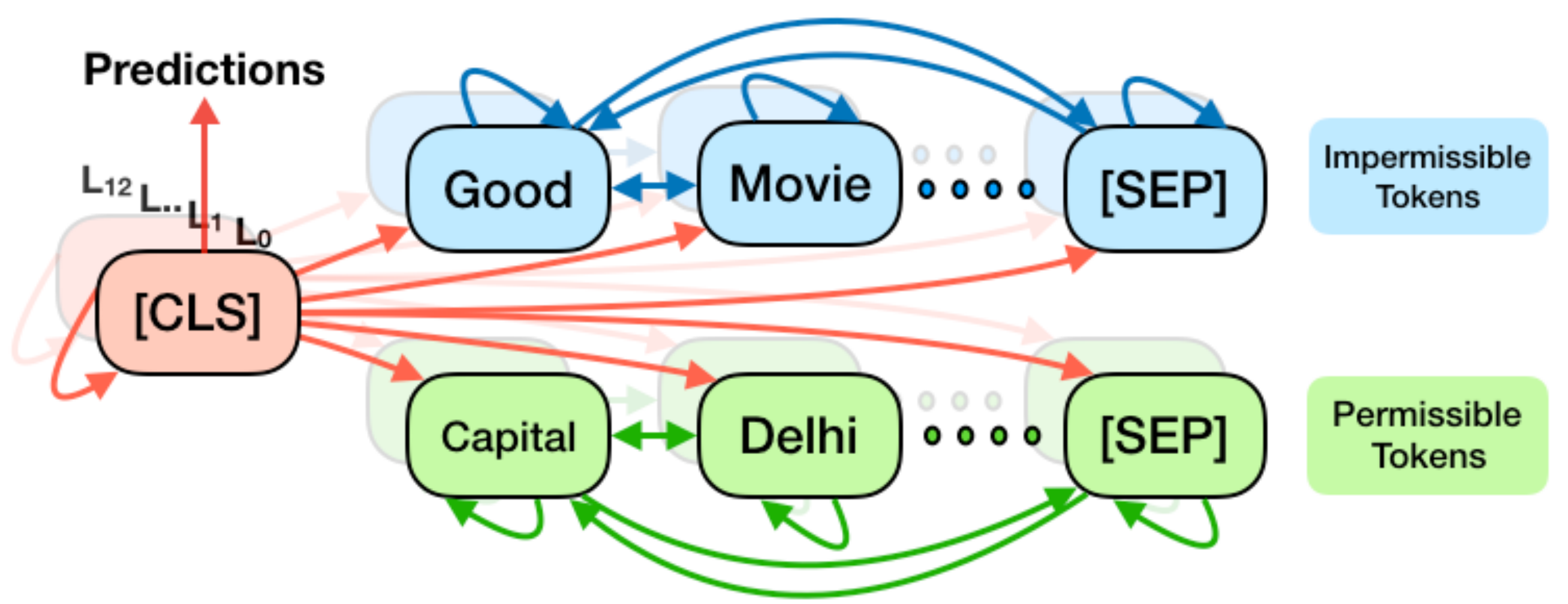}
    \caption{Restricted self-attention in BERT. The information flow through attention is restricted between impermissible and permissible tokens for every encoder layer. 
    The arrows represent the direction of attention.}
    \label{fig:flow}
    \vspace{-0.15in}
\end{figure}

\paragraph{Transformer Models} 
We use the Bidirectional Encoder Representations 
from Transformers (BERT) model \cite{devlin2018bert}.
We use the base version consisting of 12 layers with self-attention. 
Further, each of the self-attention layers consists of 12 attention heads.
The first token of every sequence is the special classification token \texttt{[CLS]}, 
whose final hidden state is used for classification tasks. 
To block the information flow from permissible to impermissible tokens, 
we multiply attention weights at every layer with a \emph{self-attention mask} $\mathbf{M}$, 
a binary matrix of size $n \times n$ where $n$ is the size of the input sequence. 
An element $\mathbf{M}_{i,j}$ represents whether the token $w_i$ 
should attend on the token $w_j$. $\mathbf{M}_{i,j}$ is $1$ 
if both $i$ and $j$ belong to the same set 
(either the set of impermissible tokens, $\mathcal{I}$ or its complement $\mathcal{I}^c$). 
Additionally,  the \texttt{[CLS]} token attends to all the tokens, 
but no token attends to \texttt{[CLS]} to prevent the information flow 
between $\mathcal{I}$ and  $ \mathcal{I}^c$ (Figure \ref{fig:flow} illustrates this setting). 
We attempt to manipulate attention from \texttt{[CLS]} token
to other tokens, and consider two variants: 
one where we manipulate the maximum attention across all heads,
and one where we manipulate the mean attention.

\subsection{Sequence-to-sequence Tasks}
Previous studies analysing the interpretability
of attention are all restricted to classification tasks
\citep{jain2019analysis,smith2019attention,pinter2019attention}.
Whereas,
attention mechanism was first introduced for,
and reportedly leads to significant gains in,
sequence-to-sequence tasks. 
Here, we analyse whether for such tasks attention can be
manipulated away from its usual interpretation as an alignment
between output and input tokens.
We begin with three synthetic sequence-to-sequence tasks 
that involve learning simple input-to-output mappings.\footnote{These tasks have been previously used in the literature to assess the ability of RNNs 
to learn long-range reorderings and substitutions~\cite{grefenstette2015learning}.}
 

\begin{table*}[ht!]
    \small
    \centering
    \begin{tabular}{lllccccccccccc}
    \toprule
    \multirow{2}{*}{Model} & \multirow{2}{*}{$\lambda$} & \multirow{2}{*}{$\mathcal{I}$} &  \multicolumn{2}{c}{Occupation Pred.}  & & \multicolumn{2}{c}{Gender Identify}  & & \multicolumn{2}{c}{SST + Wiki}     & &  \multicolumn{2}{c}{Ref. Letters}    \\ \cline{4-5} \cline{7-8} \cline{10-11} \cline{13-14}
                            & & & Acc. & A.M. &  & Acc. & A.M. &  & Acc. & A.M. & &  Acc. & A.M. \\ 
                            \midrule
    Embedding & 0.0  & \xmark & 93.8 & - & & 66.8             & -     &         & 48.9             & - &        &       74.2             &  2.3              \\ 
    Embedding & 0.0  & \cmark & 96.3 & 51.4 & & 100             & 99.2     &         &  70.7             & 48.4 &        &       77.5             &  2.3              \\ 
    Embedding & 0.1  & \cmark & 96.2 & 4.6 & & 99.4 & 3.4     &         &  67.9             & 36.4 &        &               76.8     &  0.5              \\ 
    Embedding & 1.0 & \cmark & 96.2 & 1.3 &  & 99.2 & 0.8     &         &  48.4             & 8.7 &        &       76.9             & 0.1              \\ 
    \midrule
    
    BiLSTM & 0.0   & \xmark & 93.3 & - & & 63.3             & -     &         &  49.1             & - &        &       74.7             & -              \\
    BiLSTM & 0.0   & \cmark & 96.4 & 50.3 & & 100             & 96.8     &         &  76.9             & 77.7 &        &       77.5             & 4.9              \\ 
    BiLSTM & 0.1  & \cmark &96.4 & 0.08 & & 100 &   $< 10^{-6}$   &         &  60.6             & 0.04 &        &       76.9             &  3.9              \\ 
    BiLSTM & 1.0  & \cmark & 96.7 & $< 10^{-2}$ & & 100 & $< 10^{-6}$     &         &  61.0             & 0.07 &        &       74.2            &  $< 10^{-2}$              \\ 
    \midrule
    BERT & 0.0  & \xmark & 95.0 & - & & 72.8            & -     &         &  50.4 & - &        &       68.2	&               \\ 
    BERT (mean) & 0.0  & \cmark & 97.2 & 13.9 & & 100             & 80.8     &         &  90.8 & 59.0 &        &       74.7	& 2.6              \\ 
    BERT (mean) & 0.1  & \cmark & 97.2 &  0.01 & & 99.9 & $< 10^{-3}$     &         & 90.9 &	$ < 10^{-2}$ &        &      76.2	& $< 10^{-1}$               \\ 
    BERT (mean) & 1.0  & \cmark &97.2 & $< 10^{-3}$ & & 99.9 & $< 10^{-3}$     &         &  90.6	& $< 10^{-3}$ &        &       75.2	 & $< 10^{-2}$             \\ 
    \midrule
    BERT & 0.0  & \xmark & 95.0 & - & & 72.8            & -     &         &  50.4 & - &        &       68.2	&               \\ 
    BERT (max) & 0.0  & \cmark & 97.2 & 99.7 & & 100 & 99.7   &         &  90.8             & 96.2  &        &  74.7 & 28.9     \\ BERT (max) & 0.1  & \cmark & 97.1 & $< 10^{-3}$ & & 99.9 & $< 10^{-3}$     &         & 90.7 &	$< 10^{-2}$ &                  & 76.7 & 	0.6          \\ 
    BERT (max) & 1.0  & \cmark & 97.4 & $< 10^{-3}$ & & 99.8 & $< 10^{-4}$      &         &  90.2 &	$< 10^{-3} $ &        &      75.9 &  $< 10^{-2} $        \\ 
    \bottomrule
    \end{tabular}
    \caption{Accuracy of various classification models along with their attention mass (A.M.) on impermissible tokens $\mathcal{I}$, with varying values of the loss coefficient $\lambda$. The first row for each model class represents the case when impermissible tokens $\mathcal{I}$ for the task are deleted/anonymized. For most models, and tasks, we can severely reduce attention mass on impermissible tokens while preserving original performance ($\lambda = 0$ implies no manipulation).
    }
    \label{table:acc_vs_attn_mass}
    \end{table*}

\paragraph{Bigram Flipping} The task is to reverse the bigrams in the input
$(\{w_1, w_2 \dots w_{2n-1}, w_{2n}\} \rightarrow \{w_2, w_1, \dots w_{2n}, w_{2n-1}\})$. 
\paragraph{Sequence Copying} The task requires copying the input sequence $(\{w_1, w_2 \dots w_{n-1}, w_{n}\} \rightarrow \{w_1, w_2 \dots w_{n-1}, w_{n}\})$. 
\paragraph{Sequence Reversal} The goal here is to reverse the input sequence $(\{w_1, w_2 \dots w_{n-1}, w_{n}\} \rightarrow \{w_n, w_{n-1} \dots w_2, w_1\})$. 

The motivation for evaluating on the 
synthetic tasks is that 
for any given target token, 
we precisely know the input tokens responsible.
Thus, for these tasks, the gold alignments 
act as impermissible tokens in our setup
(which are different for each output token).
For each of the three tasks, we programmatically generate $100$K
random input training sequences (with their corresponding target sequences) of length upto $32$. 
The input and output vocabulary is fixed to a $1000$ unique tokens. 
For the task of bigram flipping, the input lengths are restricted to be even. 
We use two sets of $100$K unseen random sequences from the same distribution as the validation and test set.

\paragraph{Machine Translation (English to German)} Besides synthetic tasks, 
we also evaluate on
English to German translation.
We use the Multi30K dataset, comprising of image descriptions~\cite{elliott2016multi30k}.
Since the gold target to source word-level alignment is unavailable,
we rely on the Fast Align toolkit~\cite{dyer2013simple} to 
align target words to their source counterparts. We use these aligned words as impermissible tokens.

For all sequence-to-sequence tasks, we use an encoder-decoder architecture.
Our encoder is a bidirectional GRU, and our decoder is a unidirectional GRU, 
with dot-product 
attention over source tokens, computed at each decoding timestep.\footnote{
Implementation details: the encoder and decoder token embedding size is 256, 
the encoder and decoder hidden dimension size is 512, and the teacher forcing ratio is 0.5. We use top-1 greedy strategy to decode 
the output sequence.}
We also run ablation studies with 
(i) no attention, i.e. just using the last (or the first) hidden state of the encoder; and
(ii) uniform attention, i.e. all the source tokens are uniformly weighted.\footnote{
All data and code will be released on publication.
}





\section{Results and Discussion}
\label{sec:discussion}


\begin{table*}[ht!]
 \small
  \centering
  \begin{tabular}{llccccccccccc}
  \toprule
  \multirow{2}{*}{Attention} & \multirow{2}{*}{$\lambda$} &  \multicolumn{2}{c}{Bigram Flip}  & & \multicolumn{2}{c}{Sequence Copy}  & & \multicolumn{2}{c}{Sequence Reverse}     & &  \multicolumn{2}{c}{En $\rightarrow$ De MT}    \\ \cline{3-4} \cline{6-7} \cline{9-10} \cline{12-13}
                          & & Acc. & A.M. &  & Acc. & A.M. &  & Acc. & A.M. & &  BLEU & A.M. \\ 
                          \midrule
  Dot-Product & 0.0  & 100.0 & 94.5 & & 99.9 & 98.8     &         &  100.0             & 94.1 &        &         24.4          &  20.6             \\ 
  \midrule
  Uniform & 0.0 & 97.8 & 5.2 &  & 93.8 & 5.2     &         &  88.1             & 4.7 &        &       18.5           & 5.9              \\ 
  None & 0.0  & 96.4 & 0.0 & & 84.1 & 0.0     &         &  84.1             & 0.0 &        &               14.9     &  0.0             \\ 
  \midrule

  Manipulated & 0.1  & 99.9 & 24.4 & & 100.0 &   27.3   &         &  100             & 27.6 &        &       23.7             &  7.0              \\ 
  Manipulated & 1.0  & 99.8 & 0.03 & & 92.9 &     0.02  &         &  99.8             & 0.01 &        &       20.6            &  1.1              \\ 
  \bottomrule
  \end{tabular}
 \caption{Performance of sequence-to-sequence models and their attention mass (A.M.) on impermissible tokens $\mathcal{I}$, with varying values of the loss coefficient $\lambda$. Similar to classification tasks, we can severely reduce attention mass on impermissible tokens while retaining original performance. All values are averaged over five runs.}
  \label{table:acc_vs_attn_mass_seq2seq}
  \vspace{-0.1in}
  \end{table*}

In this section we examine how
lowering attention affects task performance
(\S~\ref{subsec:attn_mass_vs_performance}). 
We then present experiments with human participants
to quantify the deception with manipulated attention (\S~\ref{subsec:human_study}). 
Lastly, we
identify alternate workarounds 
through which models preserve task performance (\S~\ref{subsec:alternate_mechanisms}).

\subsection{Attention mass and task performance} 
\label{subsec:attn_mass_vs_performance}
For the \textbf{classification tasks},
we 
experiment with the loss coefficient $\lambda \in \{0, 0.1, 1\}$. 
In each experiment, we measure the (i) attention mass: 
the sum of attention values over the set of impermissible tokens averaged over all the examples,
and (ii)
test accuracy. 
During the course of training (i.e. after each epoch), 
we arrive at different models from which 
we choose the one whose performance 
is within $2\%$ of the original accuracy 
and provides the greatest reduction 
in attention mass on impermissible tokens. 
This is done using the development set,  
and the results on the test set from the chosen model 
are presented in Table~\ref{table:acc_vs_attn_mass}.
Across most tasks, and models, 
we find 
that our manipulation scheme severely reduces 
the attention mass on impermissible 
tokens compared to models without any manipulation (i.e. when $\lambda = 0$). 
This reduction comes at a minor, or no, decrease in task accuracy. 
Note that the models 
can not achieve performance similar to the original model (as they do),
unless they rely on the set of impermissible tokens. 
This can be seen from the gap 
between models that do not use impermissible tokens
(~$\mathcal{I}$ \xmark)
from ones that do (~$\mathcal{I}$ \cmark).


\noindent The only outlier to our 
findings is the 
SST+Wiki sentiment analysis task, 
where we observe that the manipulated Embedding and BiLSTM models 
reduce the attention mass but also lose accuracy.
We speculate that these models are under parameterized
and thus jointly reducing attention mass
and retaining original accuracy is harder.
The more expressive BERT
obtains an accuracy of over $90\%$
while reducing the maximum
attention mass over the movie review from $96.2\%$ to $10^{-3}\%$.


\noindent For \textbf{sequence-to-sequence tasks}, 
from Table~\ref{table:acc_vs_attn_mass_seq2seq},
we observe that our manipulation scheme
can similarly reduce attention mass over impermissible
alignments while preserving original performance.
To measure performance, we use token-by-token accuracy
for synthetic tasks, and BLEU score for English to German MT.
We also notice that the models 
with manipulated attention 
(i.e. deliberately misaligned)
\textit{outperform models with none or uniform attention}. 
This suggests that attention mechanisms add
value to the learning process in sequence-to-sequence tasks
which goes beyond their usual interpretation as alignments.


\subsection{Human Study}
\label{subsec:human_study}
We present three human subjects 
a series of inputs and outputs from the BiLSTM models, 
trained to predict occupation (physician or surgeon) given a short biography.\footnote{The participating subjects are graduate students, proficient in English, and unaware of our work.}
We highlight the input tokens 
as per the attention scores from three different schemes:
(i) original dot-product attention, 
(ii) adversarial attention from~\citet{pinter2019attention}, and,
(iii) our proposed attention manipulation strategy.
We ask human annotators
(Q1): Do you think that this prediction was influenced by the gender of the individual? 
Each participant answers either ``yes'' or
``no'' for a set of $50$ examples 
from each of the three attention schemes.\footnote{We shuffled the order of sets among the three participants to prevent any ordering bias. Full details of the instructions presented to the annotators are in the appendix}
After looking at $50$ examples from a given attention scheme, 
we inquire about \emph{trustworthiness} of the attention scores (Q2): 
Do you believe the highlighted tokens capture the factors that drive the models' prediction? 
They answer the question on a scale of $1$ to $4$, where $1$ denotes
that the highlighted tokens \emph{do not} determine
the models' prediction, whereas $4$ implies 
they \emph{significantly} determine the models' prediction.  
We deliberately ask participants once (towards the end) 
about the trustworthiness of attention-based explanations, 
in contrast to polling after each example, 
as it requires multiple examples to assess whether the explanations 
capture factors that are predictive.




\paragraph{Results}
We find that for the original dot-product attention, 
annotators labeled $66\%$ of predictions to be influenced by gender. 
Whereas for the other two attention schemes,
none of the predictions were marked to be influenced 
by gender (see Table~\ref{tbl:human_study}).
This is despite all three models achieving roughly
the same high accuracy ($96\%$) which relies on
gender information.
This demonstrates the efficacy of our manipulation scheme---predictions
from models biased against gender minorities are 
perceived (by human participants) as not being influenced by gender.
Further, our manipulated explanations receive a 
trustworthiness score of 2.67 (out of 4), 
only slightly lower than the score for the original explanations, 
and significantly better than the adversarial attention.
We found that the KL divergence term in training
adversarial attention (Eq.~\ref{eq:adversarial})
encourages all the attention mass to concentrate on a
single uninformative token for most examples,
and hence was deemed as less trustworthy by the
annotators (see Table~\ref{tbl:human_study},
more examples in appendix).
By contrast, our manipulation scheme only reduces 
attention mass over problematic tokens, and retains
attention over non-problematic but predictive
ones (e.g. ``medicine'') making it more believable. 
We assess agreement among annotators, 
and calculate the Fleiss' Kappa to be 0.97, suggesting almost perfect agreement.

\begin{table}
\small
\centering
\begin{tabular}{@{}ccrr@{}}
\toprule
\textbf{Attention}                    & \textbf{Example}                                                                                           & \textbf{Q1} & \textbf{Q2}\\ \midrule
Original & \begin{tabular}[c]{@{}c@{}} \hl{Ms.} \texttt{X} practices \\ \hlc[lemonchiffon]{medicine} and specializes \\ in urological surgery  \end{tabular} &\begin{tabular}[c]{@{}c@{}} 66\%\\ (yes) \end{tabular} &  3.00   \\ \midrule
\begin{tabular}[c]{@{}c@{}} Adversarial\\(Wiegreffe  and\\ Pinter, 2019)\end{tabular} & \begin{tabular}[c]{@{}c@{}} Ms. \texttt{X} practices \\ medicine  and specializes \\ \hl{in}  urological surgery  \end{tabular} & \begin{tabular}[c]{@{}c@{}} 0\%\\ (yes) \end{tabular} &  1.00   \\ \midrule
Ours & \begin{tabular}[c]{@{}c@{}} Ms. \texttt{X} practices \\ \hlc[lemonchiffon]{medicine} and \hlc[lemonchiffon]{specializes} \\ in urological surgery  \end{tabular} & \begin{tabular}[c]{@{}c@{}} 0\%\\ (yes) \end{tabular} &  2.67   \\ \midrule
\end{tabular}
\caption{Results to questions posed to human participants. Q1: Do you think that this prediction was influenced by the gender of the individual? Q2: Do you believe the highlighted tokens capture the factors that drive the models’ prediction? See \S~\ref{subsec:human_study} for discussion.}
\label{tbl:human_study}
\vspace{-0.1in}
\end{table}

%

\begin{figure*}[ht]
  \centering
  \begin{subfigure}{\textwidth}
  \centering
      \begin{subfigure}{0.22\textwidth}
        \raggedright
        \includegraphics[width=\textwidth]{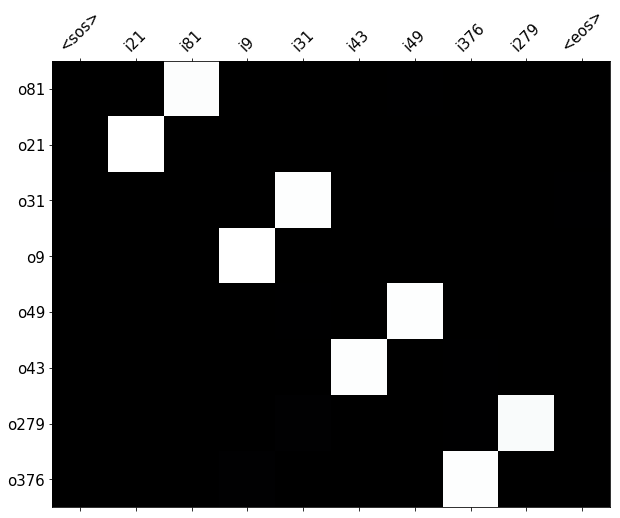}
      \end{subfigure}\qquad
      \begin{subfigure}{0.22\textwidth}
        \centering
        \includegraphics[width=\textwidth]{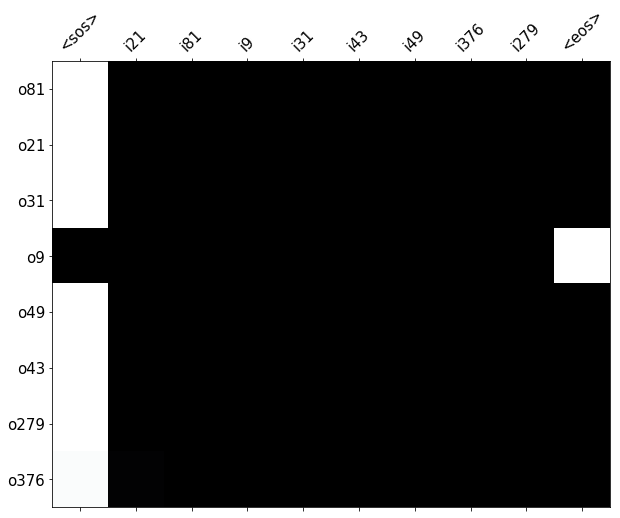}
      \end{subfigure}\qquad
      \begin{subfigure}{0.22\textwidth}
        \raggedleft
        \includegraphics[width=\textwidth]{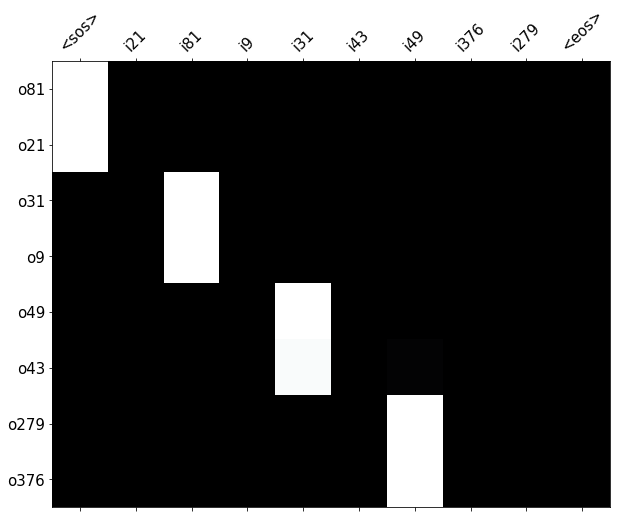}
      \end{subfigure}%
      \caption{Bigram Flipping}
\end{subfigure}

  \begin{subfigure}{\textwidth}
\centering
  \begin{subfigure}{0.22\textwidth}
    \raggedright
    \includegraphics[width=\textwidth]{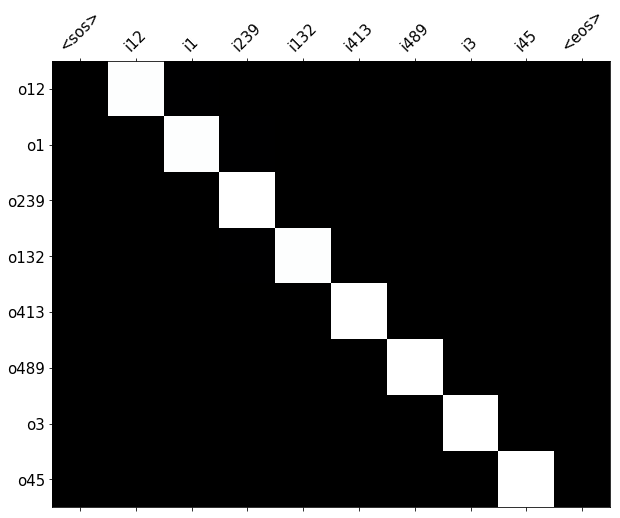}
  \end{subfigure}\qquad
  \begin{subfigure}{0.22\textwidth}
    \centering
    \includegraphics[width=\textwidth]{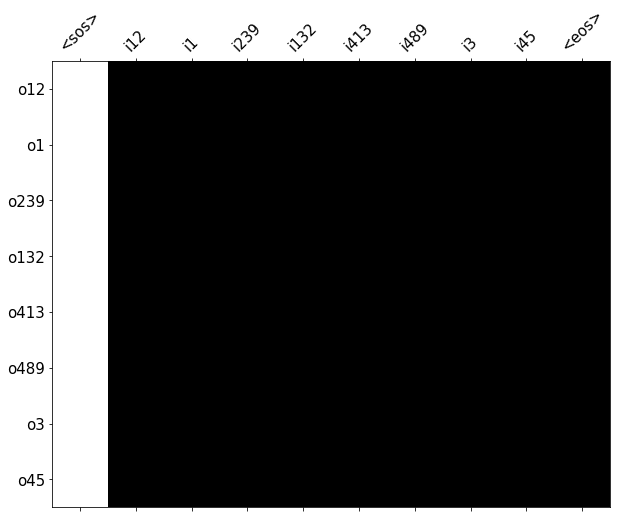}
  \end{subfigure}\qquad
  \begin{subfigure}{0.22\textwidth}
    \raggedleft
    \includegraphics[width=\textwidth]{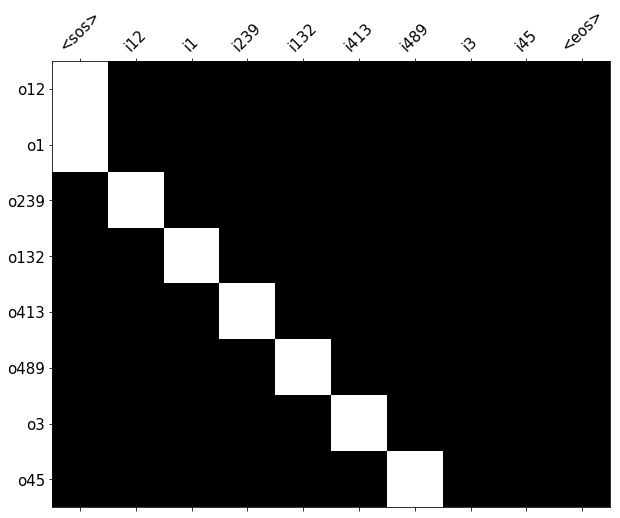}
  \end{subfigure}%
\caption{Sequence Copying}
\end{subfigure}

  \begin{subfigure}{\textwidth}
\centering
  \begin{subfigure}{0.22\textwidth}
    \raggedright
    \includegraphics[width=\textwidth]{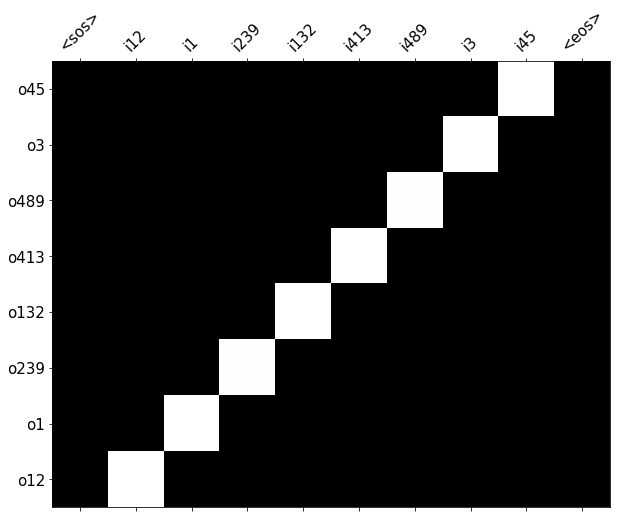}
  \end{subfigure}\qquad
  \begin{subfigure}{0.22\textwidth}
    \centering
    \includegraphics[width=\textwidth]{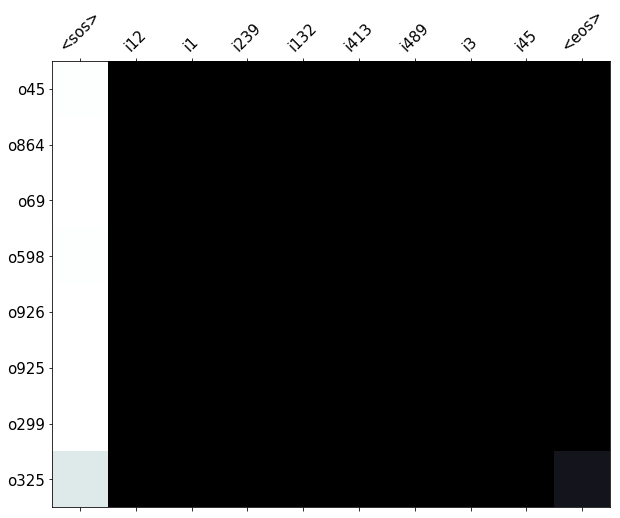}
  \end{subfigure}\qquad
  \begin{subfigure}{0.22\textwidth}
    \raggedleft
    \includegraphics[width=\textwidth]{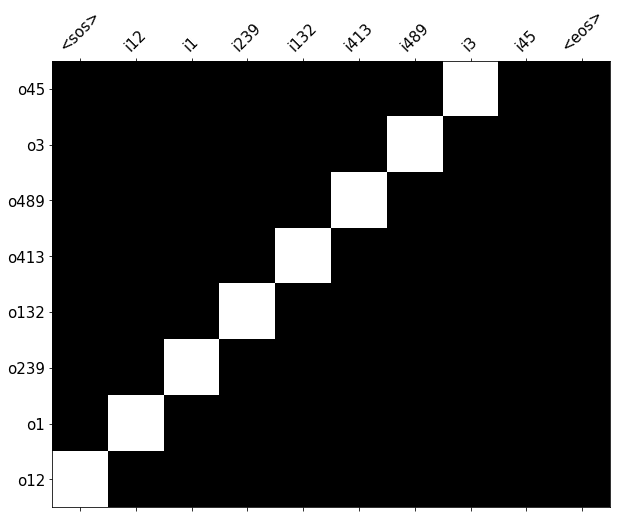}
  \end{subfigure}%
\caption{Sequence Reversal}
\end{subfigure}
\caption{For three sequence-to-sequence tasks, 
we plot the original attention map on the left, 
followed by the attention plots of two manipulated models. 
The only difference between the manipulated models for 
each task is the (random) initialization seed. 
Different manipulated models resort to different alternative mechanisms.}
\label{fig:alternative_mechanism_seq2seq}
\vspace{-0.1in}
\end{figure*}

\begin{figure}[t]
  \centering
  \includegraphics[width=.95\linewidth]{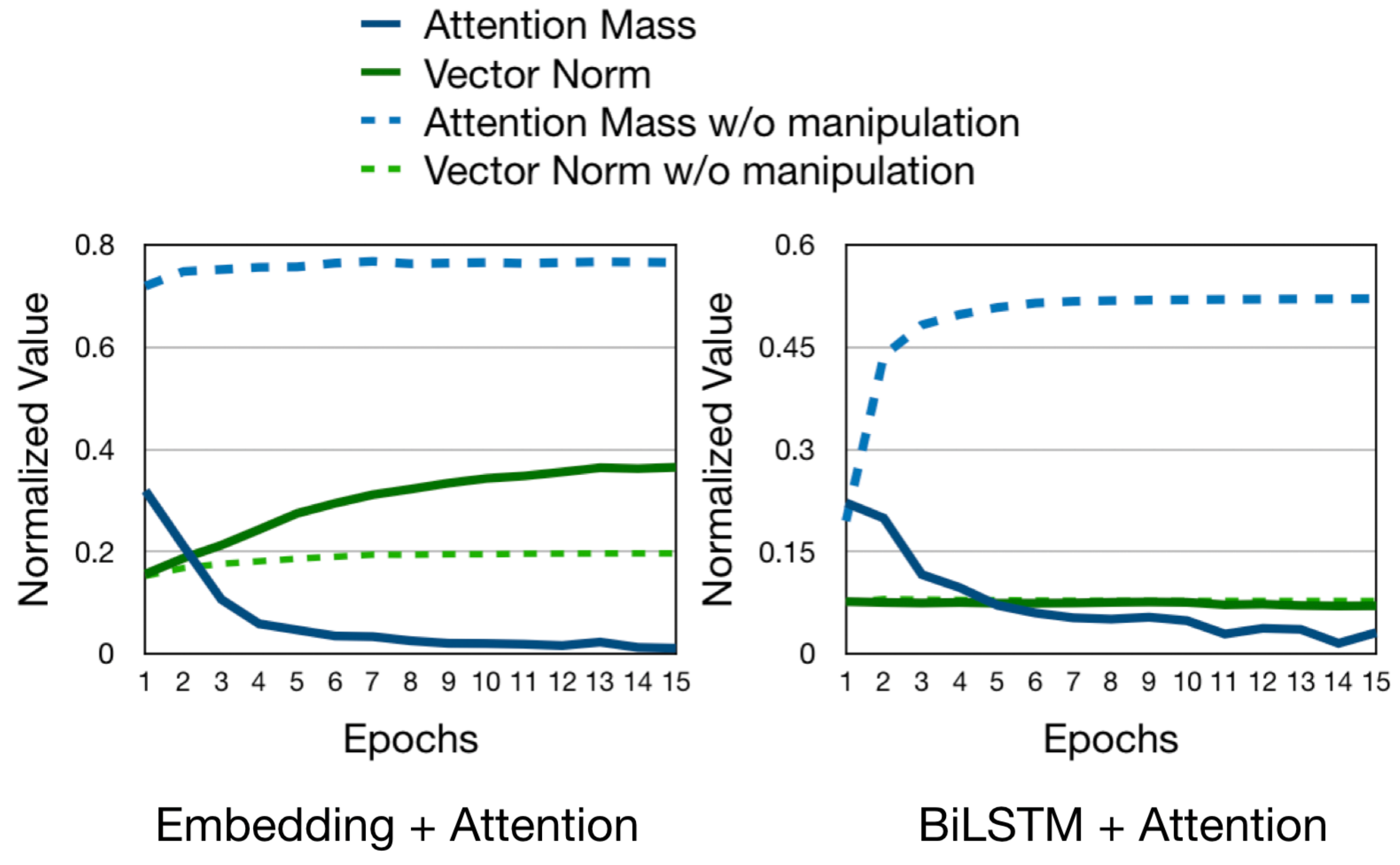}
  \caption{
  For gender identification task, the norms of embedding vectors corresponding to impermissible tokens increase considerably in Embedding+Attention model to offset the low attention values.
This is not the case for BiLSTM+Attention model as it can pass information due to recurrent connections.}
\label{fig:norms_vs_attention_mass}
\vspace{-0.15in}
\end{figure}

\subsection{Alternative Workarounds} 
\label{subsec:alternate_mechanisms}
We identify two
mechanisms by which the models \emph{cheat},
obtaining low attention values while remaining accurate. 

\textbf{Models with recurrent encoders} 
can simply pass information across tokens 
through recurrent connections, 
prior to the application of attention.
To measure this effect, 
we hard-set the attention values corresponding to 
impermissible words to zero 
\emph{after} the manipulated model is trained, thus clipping their direct contributions for inference. 
For gender classification using the BiLSTM model, 
we are still able to predict over $99\%$ of instances correctly, 
thus confirming a large degree of information flow to neighboring representations.\footnote{
A recent study~\cite{brunner2019validity} similarly observes a high degree of `mixing' of information across layers in Transformer models. 
}
In contrast, the Embedding model 
(which has no means to pass the information pre-attention) 
attains only about $50\%$ test accuracy 
after zeroing the attention values for gender pronouns. 
We see similar evidence of passing around information in sequence-to-sequence models,
where certain manipulated
attention maps are off by one or two positions 
from the gold alignments (see Figure~\ref{fig:alternative_mechanism_seq2seq}).

\textbf{Models restricted from passing information} 
prior to the attention mechanism
tend to increase the magnitude of the representations corresponding to impermissible words 
to compensate for the low attention values. 
This effect is illustrated in Figure \ref{fig:norms_vs_attention_mass}, where 
the L2 norm of embeddings for impermissible tokens increase considerably for the Embedding model during training.
We do not see increased embedding norms for the BiLSTM model,
as this is unnecessary due to the
model's capability to
move around relevant information.

We also notice that 
\textbf{differently initialized models 
attain different alternative mechanisms}. 
In Figure~\ref{fig:alternative_mechanism_seq2seq}, 
we present attention maps from the original model, 
alongside two manipulated models initialized with different seeds.
In some cases, the attention map is off by one or two positions
from the gold alignments. 
In other cases, all the attention
is confined to the first hidden state.
In such cases, manipulated models 
are similar to a
no-attention model, yet they 
offer better performance.
In preliminary experiments, we 
found a few such models
that outperform the no-attention baseline, 
even when the attention is 
turned off during inference.
This suggests
that attention offers benefits during training,
even if it is not used during inference.



\section{Conclusion}
\label{sec:conclusion}
Amidst practices that 
perceive attention scores to
be an
indication of \emph{what the model focuses on},
we 
show
that attention scores
are easily 
manipulable. 
Our simple training scheme
produces models with
significantly reduced attention mass over
tokens known a priori to be useful 
for prediction, 
while continuing to
use
them.
Our results 
raise
concerns about 
the potential use of attention
as a tool 
to audit 
algorithms, as malicious actors
could employ similar techniques to 
mislead regulators.



\section*{Acknowledgement}
The authors thank Dr. Julian McAuley for providing, and painstakingly anonymizing the data for reference letters. 
We also acknowledge Alankar Jain for carefully reading the manuscript and providing useful feedback. 
ZL thanks Amazon AI, NVIDIA, Salesforce, Facebook AI, AbridgeAI, UPMC, the Center for Machine Learning in Health, the PwC Center, the AI Ethics and Governance Fund, and DARPA's Learning with Less Labels Initiative, for their support of ACMI Lab's research on robust and societally aligned machine learning.


\bibliography{acl2020}
\bibliographystyle{acl_natbib}

\clearpage
\appendix
\section*{Supplementary Material}

\label{appendix:human_study}
\section{Instructions for human study}
In a series of examples, we present the inputs and outputs of a machine learning (ML) model trained to predict \textbf{occupation} (physician or surgeon) given a short \textbf{bio} (text). In each bio, we attempt to explain the predictions of the model. Specifically, we employ a technique that highlights words that (per our explanation method) are thought to be responsible for a particular prediction (colloquially, \emph{what the model focuses on}). 
For each unique example below, answer the following question: Do you think that this prediction was influenced by the gender of the individual? 

\begin{itemize}
    \item Yes, I suspect that the gender influenced the prediction.
    \item No, I have no reason to suspect that gender influenced the prediction.
\end{itemize}

Please note that, all the examples in this file are input, output pairs from one specific model. Further, darker shades of highlighting indicate a higher emphasis for the token (as per our explanation method). 

After showing $50$ examples from a given attention scheme, we inquire: Overall, do you believe the highlighted tokens capture the factors that drive the models’ prediction?

\begin{enumerate}
    \item The highlighted tokens capture factors that \textbf{do not} determine the models’ prediction.
    \item The highlighted tokens capture factors that \textbf{marginally} determine the models’ prediction.
    \item The highlighted tokens capture factors that \textbf{moderately} determine the models’ predictions. 
    \item The highlighted tokens capture factors that \textbf{significantly} determine the models’ predictions.
\end{enumerate}

\label{appendix:dataset}
\section{Dataset Details}
Details about the datasets used for classification tasks are available in Table~\ref{table:stats}. 

\begin{table}
\small
\centering
\begin{tabular}{@{}cccc@{}}
\toprule
\begin{tabular}[l]{@{}c@{}} \textbf{Dataset} \\ (Task) \end{tabular}                    & \textbf{Train}                                                                                            & \textbf{Val} & \textbf{Test} \\ \toprule
\begin{tabular}[c]{@{}c@{}}CommonCrawl Biographies \\ (Physician vs Surgeon) \end{tabular} &  $17629$ & $2519$ &  $5037$ \\ \midrule
\begin{tabular}[c]{@{}c@{}}Wikipedia Biographies \\ (Gender Identification) \end{tabular} & $9017$ & $1127$ & $1127$ \\ \midrule
\begin{tabular}[l]{@{}c@{}} SST + Wikipedia \\ (Sentiment Analysis) \end{tabular}                    & $6920$ & $872$ & $1821$ \\ \midrule
\begin{tabular}[l]{@{}c@{}}Reference Letters \\ (Acceptance Prediction) \end{tabular}   & $32800$ & $4097$ & $4094$  \\ \bottomrule
\end{tabular}
\caption{Number of training, validation, and test examples in various datasets used for classification tasks. 
}
\label{table:stats}
\end{table}

\label{appendix:examples}
\section{Qualitative Examples}
A few qualitative examples illustrating three different attention schemes are listed in Table~\ref{tbl:examples}.

\begin{table*}
\small
\centering
\begin{tabular}{@{}ccc@{}}
\toprule
\textbf{Attention}                    & \textbf{Input Example}                                                                                            & \textbf{Prediction} \\ \toprule
Original & \begin{tabular}[c]{@{}c@{}} \hl{Ms.} \texttt{X} practices  \hlc[lemonchiffon]{medicine} and specializes  in urological surgery  \end{tabular} & Physician    \\ \midrule
\begin{tabular}[c]{@{}c@{}} Adversarial\\(Wiegreffe  and\\ Pinter, 2019)\end{tabular} & \begin{tabular}[c]{@{}c@{}} Ms. \texttt{X} practices medicine  and specializes  \hl{in}  urological surgery  \end{tabular} & Physician  \\ \midrule
Ours & \begin{tabular}[c]{@{}c@{}} Ms. \texttt{X} practices \hlc[lemonchiffon]{medicine} and \hlc[lemonchiffon]{specializes} in urological surgery  \end{tabular} & Physician  \\ \midrule
& &  \\ \midrule
Original & \begin{tabular}[c]{@{}c@{}} \hlc[lemonchiffon]{Ms.} \texttt{X} \hl{practices} \hl{medicine} in Fort Myers, FL and specializes in family medicine \end{tabular} & Physician    \\ \midrule
\begin{tabular}[c]{@{}c@{}} Adversarial\\(Wiegreffe  and\\ Pinter, 2019)\end{tabular} & \begin{tabular}[c]{@{}c@{}} Ms. \texttt{X} practices medicine \hl{in} Fort Myers, FL and specializes in family medicine  \end{tabular} & Physician  \\ \midrule
Ours & \begin{tabular}[c]{@{}c@{}} Ms. \texttt{X} practices \hlc[lemonchiffon]{medicine} in Fort Myers, FL and \hlc[lemonchiffon]{specializes} in \hlc[lemonchiffon]{family} medicine  \end{tabular} & Physician  \\ \midrule
& &  \\ \midrule
Original & \begin{tabular}[c]{@{}c@{}} Having started his \hll{surgical} career as a general \hll{orthopaedic} surgeon, \\ Mr \texttt{X} retains a broad \hll{practice} which includes knee and hand \hll{surgery}. \\ He still does regular trauma on-call for the North Hampshire hospital \\ and treats all types of \hll{orthopaedic} problems and trauma.  \end{tabular} & Surgeon    \\ \midrule
\begin{tabular}[c]{@{}c@{}} Adversarial\\(Wiegreffe  and\\ Pinter, 2019)\end{tabular} & \begin{tabular}[c]{@{}c@{}} Having started his surgical career as a general orthopaedic surgeon, \\ Mr \texttt{X} retains a broad practice which includes knee and hand surgery. \\ He still does regular trauma on-call for \hl{the} North Hampshire hospital \\ and treats all types of orthopaedic problems and trauma.  \end{tabular} & Surgeon  \\ \midrule
Ours & \begin{tabular}[c]{@{}c@{}} Having started his \hll{surgical} career as a general \hll{orthopaedic} surgeon, \\ Mr \texttt{X} retains a broad practice which includes knee and hand surgery. \\ He still does regular trauma on-call for the North Hampshire hospital \\ and treats all types of \hll{orthopaedic} problems and trauma.   \end{tabular} & Surgeon  \\ \midrule
& &  \\ \midrule
Original & \begin{tabular}[c]{@{}c@{}} \hll{Ms.} \texttt{X} practices \hll{medicine} in ... and specializes in pediatrics. \hll{Ms.} \texttt{X} is affiliated \\ with childrens of Alabama, Saint Vincents hospital Birmingham and \\ Brookwood Medical Center. \hll{Ms.} \texttt{X} speaks English and Arabic. \end{tabular} & Physician    \\ \midrule
\begin{tabular}[c]{@{}c@{}} Adversarial\\(Wiegreffe  and\\ Pinter, 2019)\end{tabular} & \begin{tabular}[c]{@{}c@{}}Ms. \texttt{X} practices medicine \hl{in} ... and specializes in pediatrics.  Ms. \texttt{X} is affiliated \\ with childrens of Alabama, Saint Vincents hospital Birmingham and \\ Brookwood Medical Center. Ms. \texttt{X} speaks English and Arabic.   \end{tabular} & Physician  \\ \midrule
Ours & \begin{tabular}[c]{@{}c@{}} Ms. \texttt{X} practices \hll{medicine} in ... and \hll{specializes} in \hll{pediatrics}.  Ms. \texttt{X} is affiliated\\  with childrens of Alabama, Saint Vincents hospital Birmingham and \\ Brookwood Medical Center.ÒÒ Ms. \texttt{X} \hll{speaks} \hll{English} and Arabic.   \end{tabular} & Physician  \\ 
     \\ \bottomrule
\end{tabular}
\caption{Qualitative examples.
}
\label{tbl:examples}
\end{table*}

\end{document}